\pdfoutput=1

\documentclass[11pt]{article}

\usepackage{acl}

\usepackage{times}
\usepackage{latexsym}
\usepackage{amsmath}
\usepackage{amsfonts}
\usepackage{algpseudocode}
\usepackage{graphicx}

\usepackage[ruled,linesnumbered, noend]{algorithm2e}
\RestyleAlgo{ruled}

\usepackage[T1]{fontenc}

\usepackage[utf8]{inputenc}

\usepackage{microtype}

\usepackage{inconsolata}

\usepackage{subcaption}

\usepackage[inline, shortlabels]{enumitem}
\usepackage{wrapfig}

\usepackage{todonotes}

\title{BERTrend: Neural Topic Modeling for Emerging Trends Detection}

\author{Allaa Boutaleb \\
  Sorbonne University | RTE France \\ 
  \texttt{mohamed\_allaa\_eddine.boutaleb@etu.sorbonne-universite.fr} \\ 
  \AND Jérôme Picault \\
  RTE France \\
  \texttt{jerome.picault@rte-france.com} 
  \And Guillaume Grosjean \\
  RTE France \\
  \texttt{guillaume.grosjean@rte-france.com} }

\begin{document}
\maketitle
\begin{abstract}

Detecting and tracking emerging trends and weak signals in large, evolving text corpora is vital for applications such as monitoring scientific literature, managing brand reputation, surveilling critical infrastructure {\color{Black} and more generally to any kind of text-based event detection.}
Existing solutions often fail to capture the nuanced context or dynamically track evolving patterns over time. BERTrend, a novel method, addresses these limitations using neural topic modeling in an online setting. It introduces a new metric to quantify topic popularity over time by considering both the number of documents and update frequency. This metric classifies topics as noise, weak, or strong signals, flagging emerging, rapidly growing topics for further investigation. Experimentation on two large real-world datasets demonstrates BERTrend's ability to accurately detect and track meaningful weak signals while filtering out noise, offering a comprehensive solution for monitoring emerging trends in large-scale, evolving text corpora. {\color{Black}The method can also be used for retrospective analysis of past events. In addition, the use of Large Language Models together with BERTrend offers efficient means for the interpretability of trends of events.}

\end{abstract}


\section{Introduction}


The concept of weak signals, introduced by \citet{ansoff1975managing}, refers to early indicators of emerging trends that can have significant implications across various domains. These include {\color{Black} events like } shifts in public opinion in social trends, early disruptive technologies in innovation, changes in activist groups and public sentiment in politics, and potential disease outbreaks in healthcare. Monitoring and analyzing weak signals offers valuable insights for organizations, researchers, and decision-makers, aiding in informed decision-making.

Key data sources for identifying these trends include large text corpora such as news, social media, research and technology journals or reports. 
The challenges are: distinguishing meaningful weak signals from irrelevant noise, dealing with context ambiguity, and tracking the extended period over which weak signals may gain significance.

With advances in NLP and AI, researchers have developed various techniques to detect weak signals across different fields,
including statistics-based methods, graph theory, machine learning, semantic-based approaches, and expert knowledge. However, most solutions fall short in fully addressing the challenge of detecting emerging trends {\color{Black}\cite{rousseau2021}}, either by relying solely on keyword-based analysis, which misses contextual nuances, or by being static and unable to dynamically track evolving weak signals.

In this work, we introduce BERTrend, a novel framework for detecting and monitoring emerging trends and weak signals in large, evolving text corpora. BERTrend leverages neural topic modeling, specifically BERTopic, in an online learning setting to identify and track topic evolution over time. Its key contribution lies in dynamically classifying topics as noise, weak signals, or strong signals based on their popularity trends. The proposed metric quantifies topic popularity over time by considering both the number of documents within the topic and its update frequency, incorporating an exponentially growing decay if no updates occur for an extended period. By combining neural topic modeling with a dynamic popularity metric and adaptive classification thresholds, BERTrend provides a comprehensive solution for detecting and monitoring emerging trends in large-scale, evolving text corpora. {\color{Black}We discuss the qualitative results on two comprehensive datasets, including the overall evolution of trends and specific case studies. Combined with Large Language Models (LLMs), the method an efficient way of interpreting the detected trends of events through various dimensions indicating how they evolve over time.}


\section{Background}
\label{sec:background}


{\color{Black} Among past works about weak signals detection, many are \emph{keyword-based}. Thus, portfolio maps, pioneered by \citet{yoon2012detecting}, }
 involves constructing keyword emergence maps (KEM) and keyword issue maps (KIM) based on two key metrics: degree of visibility (DoV) that quantifies the frequency of a keyword within a document set; and degree of diffusion (DoD) that measures the document frequency of each keyword. Weak signals are identified as keywords with low frequency but high growth potential. Numerous studies, such as \citet{park2017future}, \citet{donnelly2019application}, \citet{lee2018identification}, \citet{roh2020exploring}, \citet{yoo2018simulation}, \citet{griol2020detecting}, have extended and refined this approach with
 multi-word analysis, signal transformation analysis, and domain-specific applications. 
However KEMs and KIMs present two major drawbacks: by focusing on keywords only, they can miss the context surrounding a weak signal ; and the output is a single snapshot, which does not gives clear clues of evolution over time.

Topic modeling has emerged as a promising approach for weak signal detection, particularly in large textual datasets. 
\textcolor{black}{Unlike general topic evolution or drift analysis, which focus on tracking changes in established topics over time, our task aims to identify early indicators of emerging trends. It emphasizes the temporal behavior and growth of small, nascent topics rather than specific content changes within established ones.}
Thus, \citet{krigsholm2019applying} and \citet{kim2019horizon} apply text mining and Latent Dirichlet Allocation (LDA) \citep{blei2003},  to identify future signals in the domain of land administration and policy research databases. 
\citet{maitre2019detection} 
integrates LDA and Word2Vec to detect weak signals in weakly structured data. 
\citet{el2021end} 
introduce furthermore two functions for deep filtering: Weakness, which measures the significance, similarity, and evolution of topics using coherence, closeness centrality, and autocorrelation metrics; and Potential Warning, which further filters the terms of the previously filtered topics to identify potential weak signals. 

While traditional topic modeling methods like LDA have been useful for weak signal detection, they have notable limitations: 
it heavily relies on pre-set topic numbers and fails to benefit from the sophisticated, contextual embeddings provided by modern pre-trained models, resulting in less nuanced analysis. Additionally, it operates on a static basis, overlooking the crucial temporal dynamics of weak signals. 
{\color{Black} RollingLDA \cite{rieger2021-rollinglda, rieger2022-dynamic} uses a rolling window for the identification of gradual topic shifts comparing topic distributions across consecutive windows, RollingLDA can detect changes in the prominence of topics over time. The fixed number of topics is a drawback. It is rather used for long-term evolution monitoring rather than detecting weak signals; interpretability of shifts is limited to keyword comparison.
}

In contrast, our approach
leverages dynamic, high-quality contextual embeddings from pre-trained models. 
Our embedding-based technique provides a richer, more adaptive analysis that does not require preset topic counts. This shift from static, keyword-based methods to dynamic, embedding-based analysis allows for a more granular and accurate tracking of the evolution and significance of weak signals over time.


\section{BERTrend}
\label{sec:algorithm}

\begin{figure*}[ht]
    \centering
    \includegraphics[width=\linewidth]{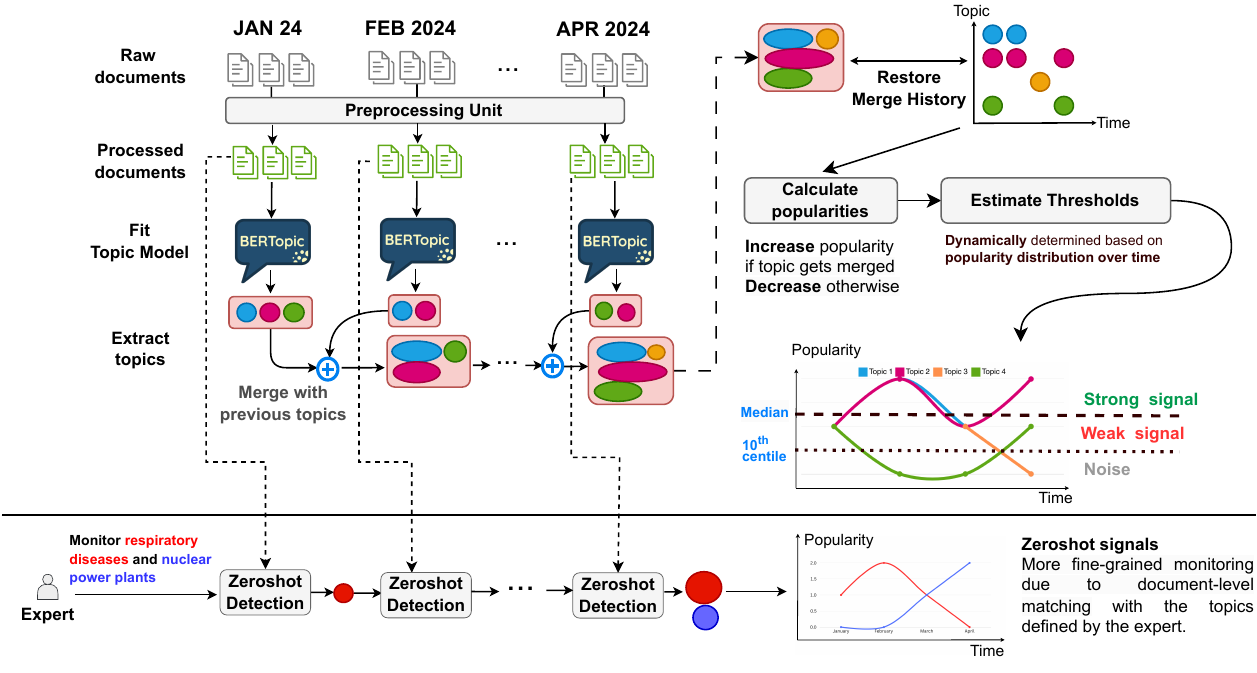}
    \caption{The BERTrend Framework processes data in time-sliced batches, undergoing preprocessing that includes unicode normalization and paragraph segmentation for very long documents. It applies a BERTopic model to extract topics for each batch, which are merged with prior batches using a similarity threshold to form a cumulative topic set. This data helps track topic popularity over time, identifying strong and weak signals based on dynamically chosen thresholds. Additionally, the framework includes a zero-shot detection feature for targeted topic monitoring, providing more fine-grained results due to document-level matching with topics defined by the expert.}
    \label{fig:algo-overview}
\end{figure*}


In this section, we describe BERTrend (Figure \ref{fig:algo-overview}), a method for identifying and tracking weak signals in large, evolving text corpora. {\color{Black}  It focuses on identifying emerging signals at a given moment, rather than tracking long-term topic evolution.} It leverages the power of BERTopic \citep{grootendorst2022bertopic}, a state-of-the-art topic model, and wraps it in an online learning framework. In this setting, new data arrives on a regular basis, allowing BERTrend to capture the dynamic evolution of topics over time. The method employs a set of metrics to characterize these topics as noise, weak signals, or strong signals based on their popularity trends. By combining the strengths of neural topic modeling with a dynamic, incremental learning approach, BERTrend enables the real-time monitoring and analysis of emerging trends and weak signals in vast, continuously growing text datasets.

BERTopic leverages pre-trained large embedding models to generate high-quality contextual embeddings of documents, enabling the discovery of meaningful and coherent topics. It utilizes HDBSCAN \citep{mcinnes2017hdbscan}, a hierarchical density-based clustering algorithm, which is robust to outliers and does not require the number of topics to be specified in advance, allowing the model to automatically determine the optimal number of topics based on the inherent structure of the data.

One of the key advantages of BERTopic is its ability to simulate online learning through model merging. Different BERTopic models can be fitted on documents from non-overlapping time periods and then merged together based on the pairwise cosine similarity between topics of consecutive models, enabling a form of dynamic topic modeling in an online learning setting.

\subsection{Data Preprocessing and Time-based Document Slicing}

To accommodate the maximum token lengths recommended by pretrained embedding models and avoid input truncation, lengthy documents are segmented into paragraphs. Each paragraph is treated as an individual document, with a mapping to its original long document source. This ensures accurate calculation of a topic's popularity over time by considering the original number of documents rather than the inflated number of paragraphs.\textcolor{black}{We filter out documents that don't contain at least 100 Latin characters. This threshold was determined by analyzing the corpus of NYT and arXiv after splitting by paragraphs. Documents below this threshold often represent noise (e.g., article endings, incomplete sentences, social media references).}

After preprocessing, the entire text corpus \(D\), consisting of \(N\) documents, is divided into document slices based on a selected time granularity (e.g., daily, weekly, monthly). A document slice \(D_t\) is defined as a subset of documents from \(D\) that fall within a specific time interval \([t, t+\Delta t)\), where \(t \in \{t_1, t_2, \ldots, t_M\}\), \(\Delta t\) is the chosen time granularity, and \(M\) is the total number of document slices. This slicing is crucial for analyzing the temporal dynamics of topics within the corpus.

\subsection{Topic Extraction using BERTopic}

For each document slice \(D_t\), BERTopic extracts a set of topics \(\mathcal{T}_t = \{\tau_t^1, \tau_t^2, \ldots, \tau_t^{K_t}\}\), where \(K_t\) is the number of topics in \(D_t\). The process involves:

1. \textit{Document Embedding}: Each document \(d \in D_t\) is transformed into a dense vector \(\mathbf{e}_d \in \mathbb{R}^h\) using a pre-trained sentence transformer model \citep{reimers2019sentence}, where \(h\) is the embedding dimension. A topic \(\tau_t^j\) is described as a set of words \(W_{\tau_t^j} = \{w_t^{j,1}, w_t^{j,2}, \ldots, w_t^{j,M_j}\}\), where \(M_j\) is the number of words representing the topic.

2. \textit{Dimensionality Reduction}: The embeddings are reduced to a lower-dimensional space using UMAP \citep{mcinnes2018umap}, resulting in reduced embeddings \(\mathbf{e}'_d \in \mathbb{R}^r\), where \(r < h\).

3. \textit{Document Clustering}: The reduced embeddings are clustered using HDBSCAN \citep{mcinnes2017hdbscan}, to group semantically similar documents into clusters. Each cluster \(\mathcal{C}_t^j \in \mathcal{C}_t\) is associated with a centroid embedding \(\mathbf{c}_t^j \in \mathbb{R}^r\). These clusters represent preliminary groupings of documents that will later be labeled as topics.


4. \textit{Cluster Labeling}: BERTopic assigns labels to clusters to form topics using class-based TF-IDF (c-TF-IDF), considering the frequency and specificity of words within each cluster. Various methods, including LLMs, KeyBERT, and Maximal Marginal Relevance (MMR), can be used to refine the representation of topics. \textcolor{black}{In our work, we maintained the default c-TF-IDF representation without employing additional refinement methods.} After labeling, each cluster $(\mathcal{C}_t^j)$ becomes a topic $(\tau_t^j)$.

\begin{algorithm}[h]
\caption{BERTrend Algorithm}
\label{alg:bertrend_algorithm}

{\small
\KwIn{Text corpus \(D\), retrospective window size \(W\), time granularity \(G\), similarity threshold \(\tau\), decay factor \(\lambda\)}
\KwOut{Topics \(\mathcal{T}\), popularity \(p\), signal categories \(S\)}

Initialize \(\mathcal{T} = \emptyset\), \(p = \emptyset\), \(S = \emptyset\)\;
\(t_{\text{now}} = \text{current time}\)\;
\(t_{\text{start}} = t_{\text{now}} - W\)\;
\(\text{time slices} = \text{slice data}(D, t_{\text{start}}, t_{\text{now}}, G)\)\;

\For{\(D_t \in \text{time slices}\)}{
  \(\mathcal{T}_t = \text{BERTopic}(D_t)\)\;
  \For{\(\tau_t^j \in \mathcal{T}_t\)}{
    \(\text{sim}_{\text{max}} = \max_{\tau_t^k \in \mathcal{T}} \text{Similarity}_{cos}(\mathbf{c}_t^j, \mathbf{c}_t^k)\)\;
    \eIf{\(\text{sim}_{\text{max}} \geq \tau\)}{
      \(k^* = \arg\max_{k} \text{Similarity}_{cos}(\mathbf{c}_t^j, \mathbf{c}_t^k)\)\;
      \(D_t^{k^*} = D_t^{k^*} \cup D_t^j\)\;
      \(p_t^{k^*} = p_{t-1}^{k^*} + |D_t^j|\)\;
    }{
      \(\mathcal{T} = \mathcal{T} \cup \{\tau_t^j\}\)\;
      \(p_t^j = |D_t^j|\)\;
    }
  }
  \For{\(\tau_t^k \in \mathcal{T}\)}{
    \If{\(\tau_t^k \notin \mathcal{T}_t\)}{
      \(p_t^k = p_{t-1}^k \cdot e^{-\lambda \Delta t^2}\)\;
    }
  }
  \(\mathbf{P}_{\text{all}} = \bigcup_{\tau^k \in \mathcal{T}} \{p_j^k \mid j \in [t-W+1, t]\}\)\;

  \(\mathbf{P}_{\text{all}} = \text{sort}(\mathbf{P}_{\text{all}})\)\;
  \(P_{10} = \mathbf{P}_{\text{all}}[\lfloor 0.1 \cdot |\mathbf{P}_{\text{all}}| \rfloor]\)\;
  \(P_{50} = \mathbf{P}_{\text{all}}[\lfloor 0.5 \cdot |\mathbf{P}_{\text{all}}| \rfloor]\)\;
  \For{\(\tau_t^k \in \mathcal{T}\)}{
    \eIf{\(p_t^k < P_{10}\)}{
      \(S_t^k = \text{"noise"}\)\;
    }{
      \eIf{\(P_{10} \leq p_t^k \leq P_{50}\)}{
        \If{\(\text{slope}(\{p_j^k \mid j \in [t-W+1, t]\}) > 0\)}{
          \(S_t^k = \text{"weak"}\)\;
        } \Else {
          \(S_t^k = \text{"noise"}\)\;
        }
      }{
        \(S_t^k = \text{"strong"}\)\;
      }
    }
  }
}
}
\end{algorithm}

\subsection{Topic Merging}

BERTrend merges topics across document slices to capture their evolution. \textcolor{black}{The topic merging process is formalized in Algorithm \ref{alg:bertrend_algorithm} (lines 10-12).} For each time-based document slice \(D_{t+1}\), the extracted topics \(\mathcal{T}_{t+1}\) are compared with the topics from the previous slice \(\mathcal{T}_t\) as follows:
\begin{enumerate}[nolistsep, leftmargin=*]
    \item \textit{Similarity Calculation}: Compute the cosine similarity between each topic embedding \(\mathbf{c}_{(t+1)}^j \in \mathcal{T}_{t+1}\) and all topic embeddings \(\mathbf{c}_t^k \in \mathcal{T}_t\).
    \item \textit{Topic Matching}: If the maximum similarity between \(\mathbf{c}_{(t+1)}^j\) and any \(\mathbf{c}_t^k\) exceeds a threshold \(\alpha\) (e.g., \(\alpha = 0.7\)), merge the topics and add the documents associated with \(\tau_{(t+1)}^j\) to \(\tau_t^k\).
     \item \textit{New Topic Creation}: If the maximum similarity is below \(\alpha\), consider \(\tau_{(t+1)}^j\) as a new topic and add it to \(\mathcal{T}_{t}\).
\end{enumerate}

To maintain topic embedding stability, the embedding of the first occurrence of a topic is retained, preventing drift and over-generalization.

\subsection{Popularity Estimation}

BERTrend estimates topic popularity over time and classifies them into signal categories based on popularity dynamics. The popularity of topic \(\tau_t^k\) for document slice \(D_t\) is denoted as \(p_t^k\) and calculated as follows:

\begin{enumerate}[nolistsep, leftmargin=*]
    \item \textit{Initial Popularity}: For a new topic \(\tau_t^k\) of document slice \(D_t\), its initial popularity is set to the number of associated documents: \(p_t^k = |D_t^k|\), where \(D_t^k\) is the set of documents associated with \(\tau_t^k\) at time \(t\).

    \item \textit{Popularity Update}: For subsequent document slices \(D_{t'}\) (\(t' > t\)):

\begin{itemize}[nolistsep, leftmargin=*]
    \item If \(\tau_t^k\) is merged with a topic in \(\mathcal{T}_{t'}\), its popularity is incremented by the number of new documents: \(p_{t'}^k = p_{t'-1}^k + |D_{t'}^k|\).
    \item If \(\tau_t^k\) is not merged with any topic in \(\mathcal{T}_{t'}\), its popularity decays exponentially: \(p_{t'}^k = p_{t'-1}^k \cdot e^{-\lambda \Delta t^2}\), where \(\lambda\) is a constant decay factor (e.g., \(\lambda = 0.01\)) and \(\Delta t\) is the number of days since \(\tau^k\) last received an update.
\end{itemize}
\end{enumerate}

\subsection{Trend Classification}

To classify topics into signal categories, BERTrend calculates percentiles of popularity values over a rolling window of size \(W\). For each document slice \(D_t\), two empirical thresholds - the 10th percentile (\(P_{10}\)) and the 50th percentile (\(P_{50}\)) of popularity values within the window \([t-W, t]\) - are computed. Trend classification is performed based on the topic's popularity \(p_t^k\) and its recent popularity trend:

\begin{itemize}[nolistsep, leftmargin=*]
    \item If \(p_t^k < P_{10}\), \(\tau_t^k\) is classified as a "noise" signal.
    \item If \(P_{10} \leq p_t^k \leq P_{50}\):
    \begin{itemize}[nolistsep, leftmargin=*]
        \item If the topic's popularity has been increasing over the past few days, as determined by a positive slope of the linear regression line fitted to the topic's popularity values within the window \([t-W, t]\), \(\tau_t^k\) is classified as a "weak" signal.
        \item If the topic's popularity has been decreasing, as determined by a negative slope of the linear regression line, \(\tau_t^k\) is classified as a "noise" signal, as it likely represents a previously popular topic that is losing relevance.
    \end{itemize}
    \item If \(p_t^k > P_{50}\), \(\tau_t^k\) is classified as a "strong" signal.
\end{itemize}

{\color{Black}BERTrend combines popularity trends with thresholds to identify emerging trends, distinguishing them from declining popular topics. This helps filter out fading "weak signals" that are actually strong but declining trends.}

Using percentiles calculated dynamically over a sliding window offers several advantages:

\begin{enumerate}[nolistsep, leftmargin=*]
    \item \textit{Adaptability to datasets}: The retrospective parameter allows the method to adapt to the input data's velocity and production frequency.
    \item \textit{Forget gate mechanism}: 
    The sliding window avoids the influence of outdated signals on current threshold calculations.
    \item \textit{Robustness to outliers}: Calculating thresholds based on the popularity distribution reduces sensitivity to outlier popularities and prevents thresholds from approaching zero when many signals have faded away.
\end{enumerate}

\subsection{Targeted Zero-shot Topic Monitoring }

BERTrend includes an optional zero-shot detection feature that allows domain experts to define a set of topics \(\mathcal{Z} = \{z_1, z_2, \ldots, z_L\}\), each represented by a textual description. The embeddings of these topics and the documents in each slice \(D_t\) are calculated using the same embedding model. For each document \(d \in D_t\), the cosine similarity between its embedding \(\mathbf{e}_d\) and the embedding of each defined topic \(z_l\) is computed. Documents with a similarity score above a predefined low threshold \(\beta\) (typically 0.4-0.6) for any of the defined topics are considered relevant and included in the corresponding topic's document set \(D_t^{z_l}\). The low threshold accounts for the presumed vagueness and generality of the expert-defined topics, as they have incomplete knowledge that would be supplemented by new emerging information. Finally, the popularity and trend classification for the zero-shot topics are performed in the same manner as for the automatically extracted topics, using the document sets \(D_t^{z_l}\) instead of \(D_t^k\).

\section{Experimental Setup}
\label{sec:experimental_setup}


\subsection{Datasets}



{\color{black}We selected two diverse datasets for our evaluation: the arXiv dataset, comprising scientific paper abstracts from the computer science category (cs.*) \citep{arxiv_dataset}, and the New York Times (NYT) news dataset \citep{nyt_dataset}. Our choice aligns with recommendations from \citet{rousseau2021} and \citet{yoon2012detecting}, who advocate for the use of scientific articles and news sources in weak signal detection due to their rich, evolving content.} The arXiv dataset spans from January 2017 to December 2023, encompassing 367,248 abstracts, while the NYT dataset covers the period from January 2019 to January 2023, including 184,811 articles. {\color{black}These corpora offer a wealth of interpretable topics, facilitating qualitative analysis and interpretation. Moreover, the NYT dataset has been previously employed in weak signal detection research \citep{el2021end}, further substantiating its relevance to our study.} These datasets were chosen for their diverse content and potential to contain topics that could be considered weak signals, such as early warnings about the COVID-19 pandemic.

\subsection{Algorithm parameters}

In our experiments, we used the BERTopic framework with carefully selected hyperparameters to optimize weak signal detection performance. We chose the "all-mpnet-base-v2"
\footnote{\url{https://huggingface.co/sentence-transformers/all-mpnet-base-v2}} 
sentence transformer for document embedding because of its strong performance on various natural language understanding tasks \citep{reimers2019sentence}.

In the UMAP dimensionality reduction step, the number of components is set to 5 (default value), and the number of neighbors to 15,
which allows UMAP to balance local and global structure in the data, as lower values focus more on local structure while higher values emphasize broader patterns \citep{mcinnes2018umap}.
In the HDBSCAN clustering step, we set the minimum cluster size to 2, the smallest possible value, to detect fine-grained clusters. 
The minimum sample size was set to 1, the smallest possible value, to reduce the likelihood of points being declared as noise, as the high number of clusters obtained reduces the need for conservative clustering \citep{mcinnes2017hdbscan}.


Topics were represented by top unigrams and bigrams based on their c-TF-IDF scores. {\color{black}To determine the optimal minimum similarity threshold for merging topics across time slices, we conducted an ablation study varying the threshold from 0.5 to 0.95. We observed that lower thresholds (0.5-0.6) led to overly broad signals and unstable behavior, characterized by a phenomenon we term "threshold collapse." In this scenario, the disproportionate merging of topics results in a few dominant signals that skew the distribution of popularity values. Consequently, the dynamically determined classification thresholds (Q1 and Q3) become volatile, potentially shifting dramatically between consecutive timestamps. This instability compromises the reliability of signal categorization.}

{\color{black}Conversely, higher thresholds (0.8-0.95) resulted in an overabundance of micro-signals, hindering the detection of meaningful trends. A threshold of 0.7 was found to provide a balanced approach, ensuring coherence and consistency of detected topics while allowing for semantic evolution without inducing threshold instability.}

{\color{black}We also investigated the effect of the retrospective window size, varying it from 2 to 30 days. We found that its impact on BERTrend's performance was minimal when using an appropriate merge similarity threshold. The choice of window size primarily depends on the desired amount of historical data to incorporate in threshold calculations, with larger windows providing more stable, but potentially less responsive, threshold determinations.}

For the granularity of the time slices, we chose 2 and 7 days for the NYT News and arXiv datasets respectively, {\color{black}based on our analysis of topic evolution rates in these datasets. This selection accommodates the rapidly evolving nature of news compared to the slower pace of research papers, while maintaining a balance between signal detection sensitivity and computational efficiency.}

{\color{black}It is important to note that these parameter choices have been fine-tuned based on the characteristics of the datasets used in this study. For datasets with significantly different topic evolution dynamics and update frequencies, these parameters may require adjustment to achieve optimal performance.}

In the zero-shot example (subsection 5.4), we used a lower similarity threshold of 0.45 for merging topics to accommodate the vague and incomplete nature of the user-defined topics, allowing for a more flexible merging process. This approach maximizes the recall in detecting potentially relevant documents of weak signals.


\section{Results}
\label{sec:results}


{\color{Black}Quantitative results about weak signal analysis are very challenging to obtain due to the lack  of established metrics and methodology as detailed in section \ref{sec:evaluation_challenges}. Therefore, as in many past works in this research area (e.g. \cite{el2021end}, we focus on a qualitative analysis, including retrospective analysis of known outcomes, }
to highlight its effectiveness and potential applications.


\subsection{Overall results}

\begin{figure*}[ht]
\centering
\begin{subfigure}{0.49\linewidth}
    \includegraphics[width=\linewidth]{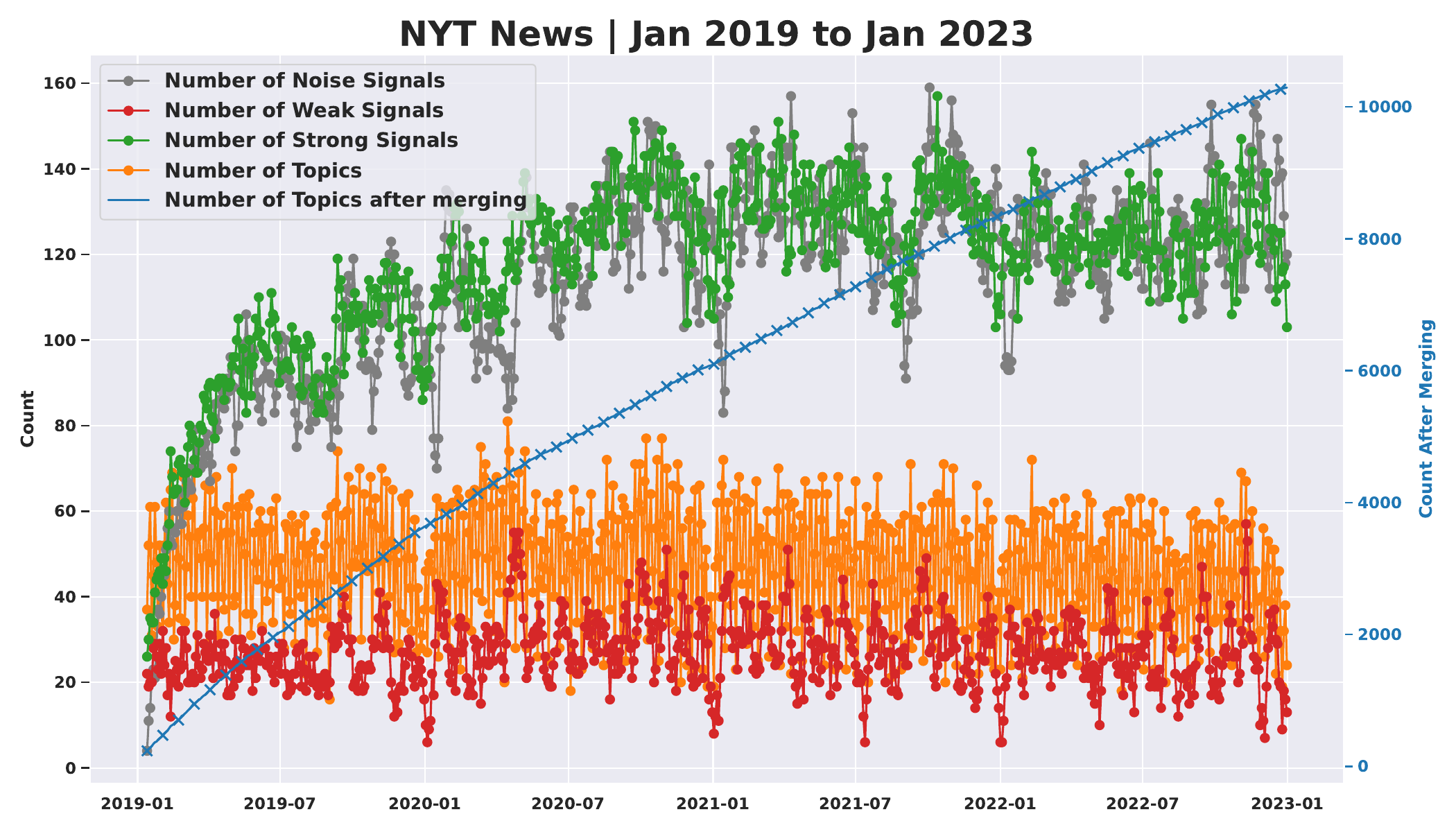}
    \caption{NYT News dataset}
\end{subfigure}
\hfill
\begin{subfigure}{0.5\linewidth}
    \includegraphics[width=\linewidth]{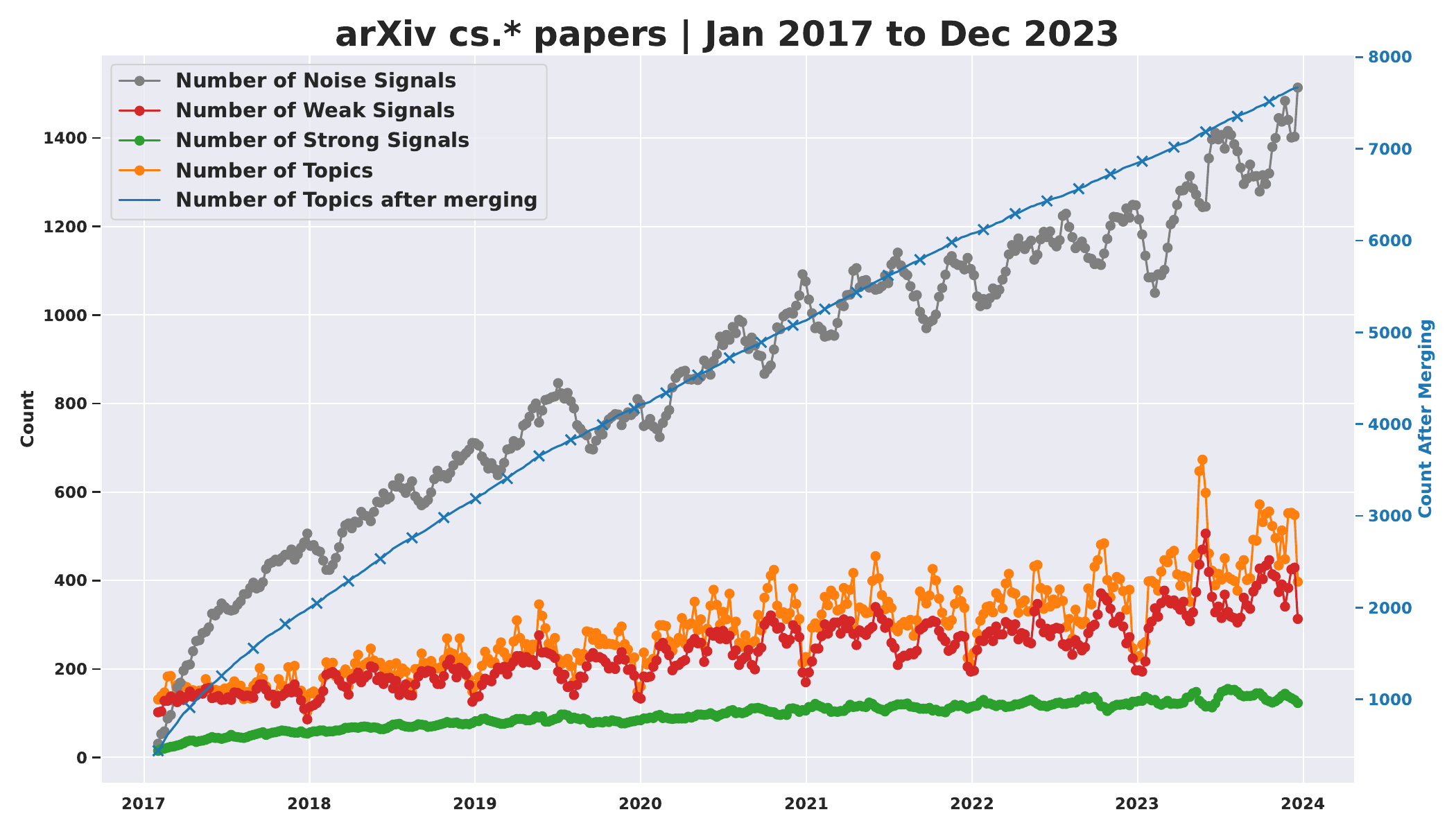}
    \caption{arXiv cs.* dataset}
\end{subfigure}
\caption{Evolution of Signal Types and Topic Counts in the NYT News and arXiv cs.* Datasets}
\label{fig:signal_evolution}
\end{figure*}

Figure \ref{fig:signal_evolution} illustrates the evolution of signal type counts and topic counts in the NYT News dataset 
and the arXiv cs.* papers dataset 
We observe striking differences in the signal type distributions between these datasets, which can be attributed to the very nature of their respective domains.

In the NYT News dataset, the number of weak signals remains relatively stable over time, with a manageable quantity of 10 to 20 signals every 2 days. This is well-suited for real-time monitoring and trend detection in fast-paced news cycles, where emerging signals quickly evolve into hot topics of discussion. The occasional spikes in strong signals likely correspond to major events or trending news stories that capture significant attention.

Conversely, the arXiv cs.* papers dataset exhibits a consistently higher number of weak signals, reflecting the diverse range of emerging research topics in the computer science domain. The number of strong signals is comparatively lower, as only a subset of novel ideas and approaches eventually gain traction and become widely adopted. This aligns with the nature of scientific research, where numerous proposals emerge, but only a few ultimately make a significant impact.

Interestingly, while the number of topics per time slice in the NYT News dataset fluctuates but remains overall stable, the arXiv cs.* papers dataset shows an increasing trend in the number of topics detected per 7-day interval. This can be attributed to the exponential growth of research papers in recent years, leading to a more diverse and rapidly evolving research landscape.
The total number of topics after merging (blue line) steadily increases over time in both datasets, reflecting the accumulation of new topics as the datasets grow. 


\subsection{Case study}

In this section, we conduct a qualitative analysis of the results. We focus on a subset of illustrative topics and zoom into key periods to observe their behavior more closely. The examples 
are selected  for their ease for interpretation.

\begin{figure*}[ht]
\centering
\begin{subfigure}{0.485\linewidth}
    \includegraphics[width=1\linewidth]{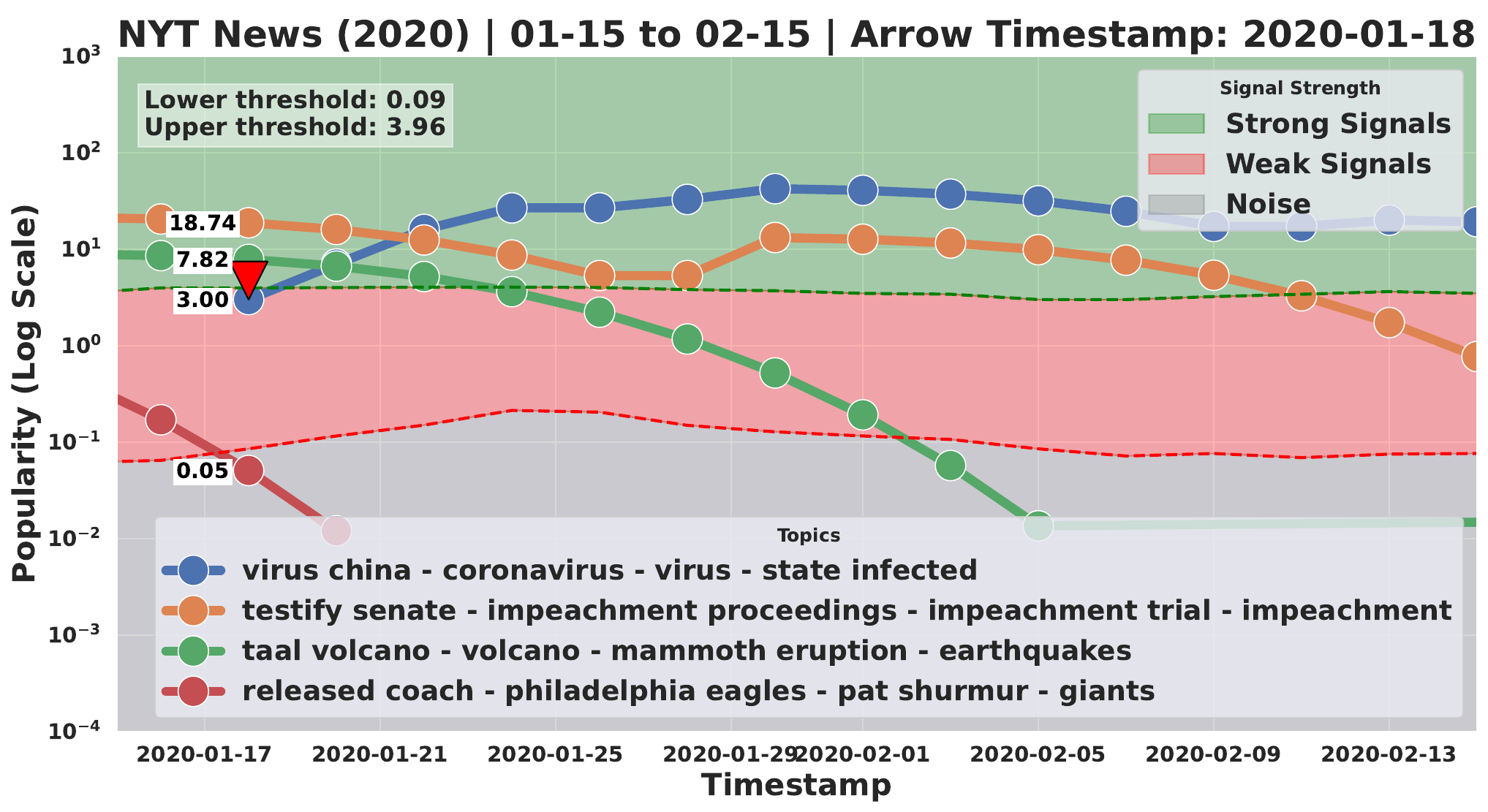}
    \caption{NYT News dataset}
\end{subfigure}
\begin{subfigure}{0.5\linewidth}
    \includegraphics[width=1\linewidth]{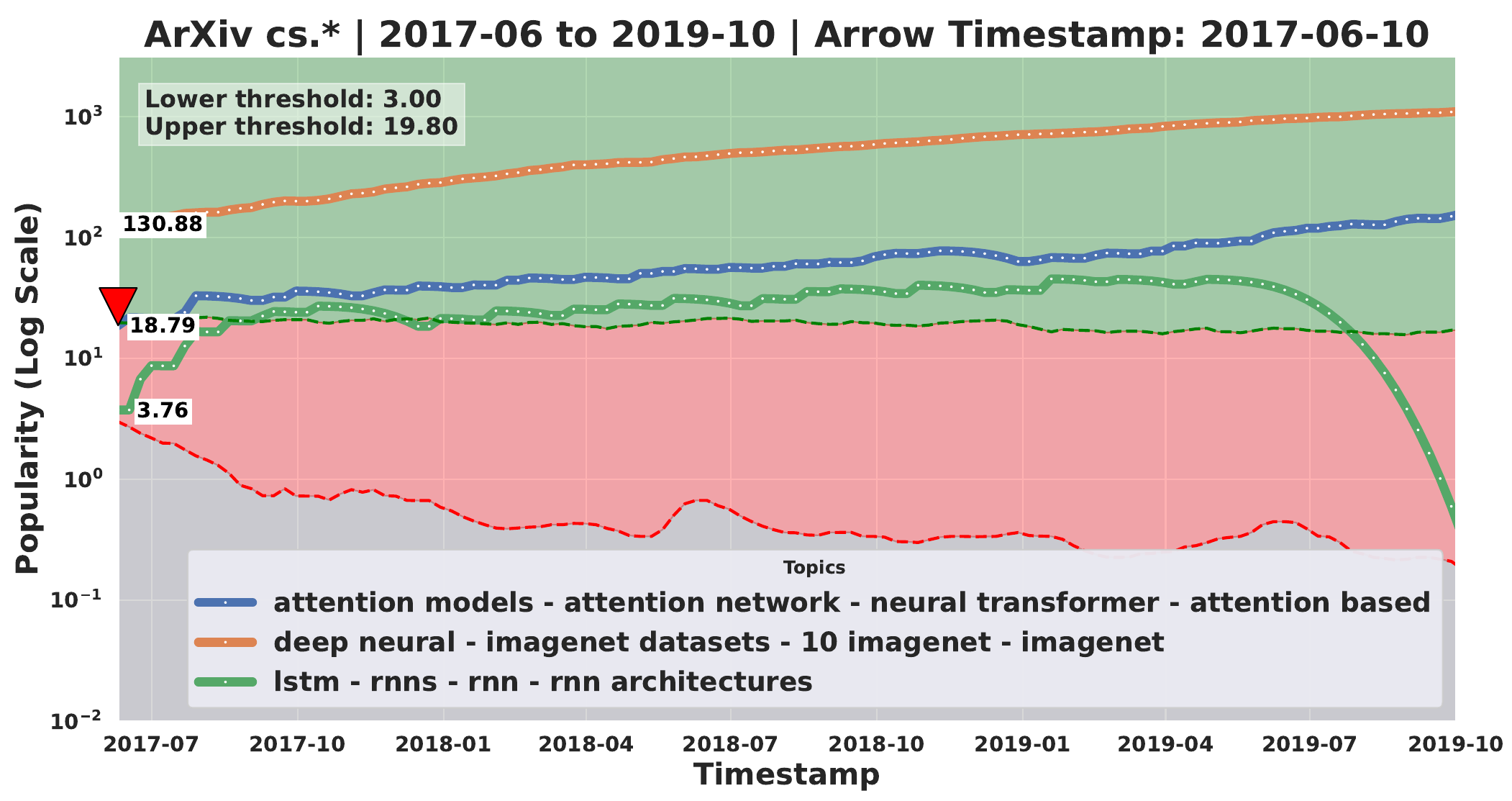}
    \caption{arXiv cs.* papers}
\end{subfigure}
\caption{Log-scaled popularity of selected topics from (a) the NYT News dataset and (b) arXiv cs.* papers.}
\label{fig:topic_popularity_examples}
\end{figure*}



Figure \ref{fig:topic_popularity_examples}a focuses on the period 
from 01/2020 to 02/2020, when news media began reporting on the COVID-19 outbreak. We observe the appearance of a new topic (blue signal), due to its dissimilarity with pre-existing topics. Initially, the blue signal is classified as  weak because of the low number of articles discussing it. Shortly after, it gains traction, transitioning from a weak to a strong signal within a matter of days, as evidenced by its exponential rise in popularity on the log-scaled y-axis. Concurrently, other strong signals during this period include topics related to the impeachment trial of President Trump (orange signal) and the Taal Volcano eruption (Philippines)  in Jan 2020 (green signal), while a topic discussing American football teams (red signal) is classified as noise.

In Figure \ref{fig:topic_popularity_examples}b, we showcase the evolution of three selected topics from the arXiv cs.* papers dataset from 06/2017 to 10/2019. The blue signal, representing attention models, was initially a weak signal before June 2017, as attention methods were being used in conjunction with recurrent networks. However, the introduction of the transformer architecture \citep{vaswani2017attention} in June 2017 marked a turning point, after which the topic quickly gained traction, transitioning into a strong signal and eventually becoming a mega-trend. This rise of transformers largely replaced RNNs \citep{rumelhart1986learning} and LSTMs \citep{hochreiter1997long} (green signal) in NLP tasks, leading to a decline in the popularity of the green signal. In contrast, papers related to computer vision, especially those mentioning ImageNet \citep{deng2009imagenet}, a widely-used dataset in computer vision, were classified as strong signals in June 2017 and continued to exhibit growth. This analysis demonstrates our method's ability to identify potentially impactful research topics early on, track their evolution, and capture the dynamics between related topics.

\subsection{Impact of zero-shot Topic Modeling}

\begin{figure}
    \centering
    \includegraphics[width=\linewidth]{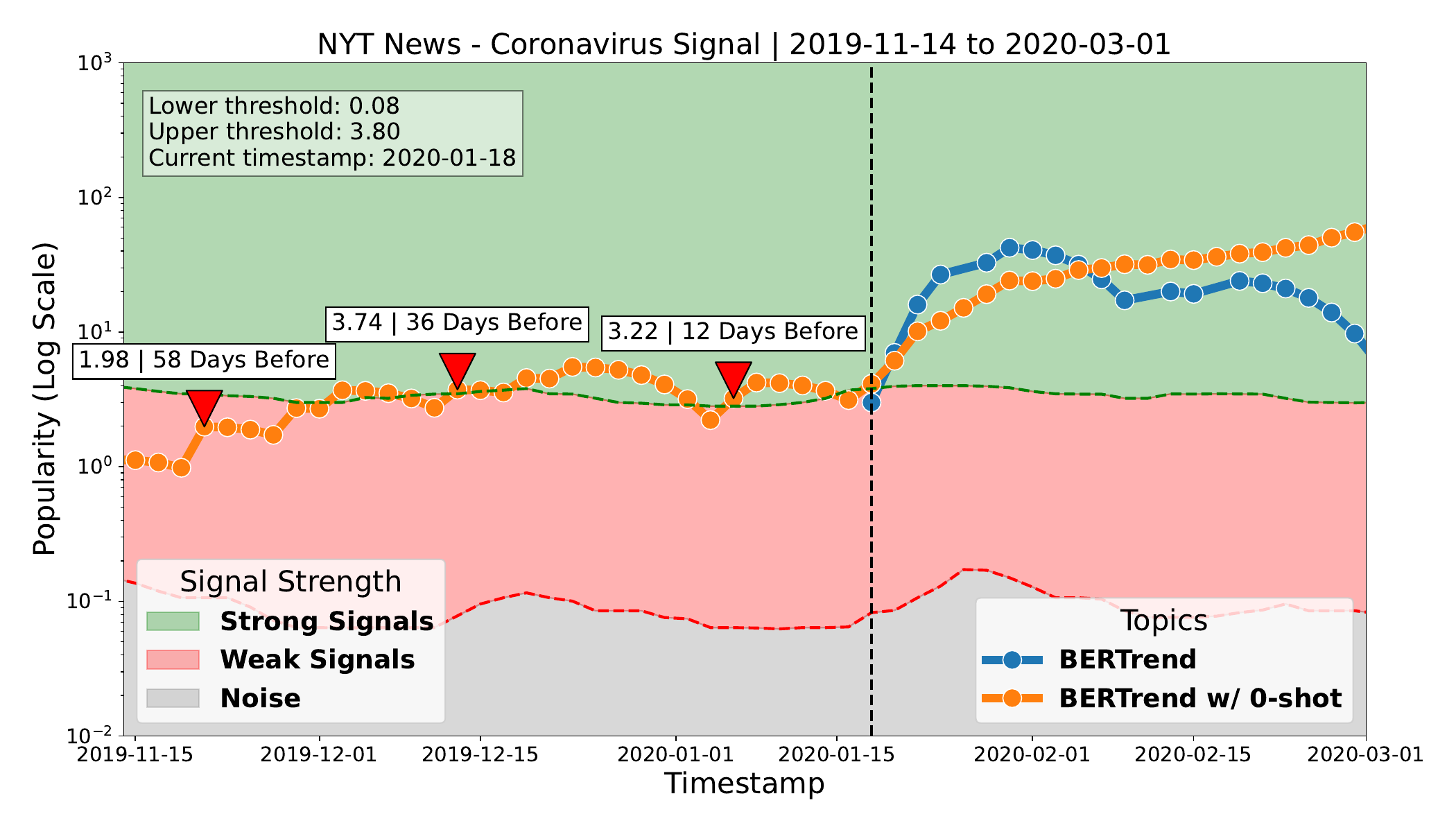}
    \includegraphics[width=\linewidth]{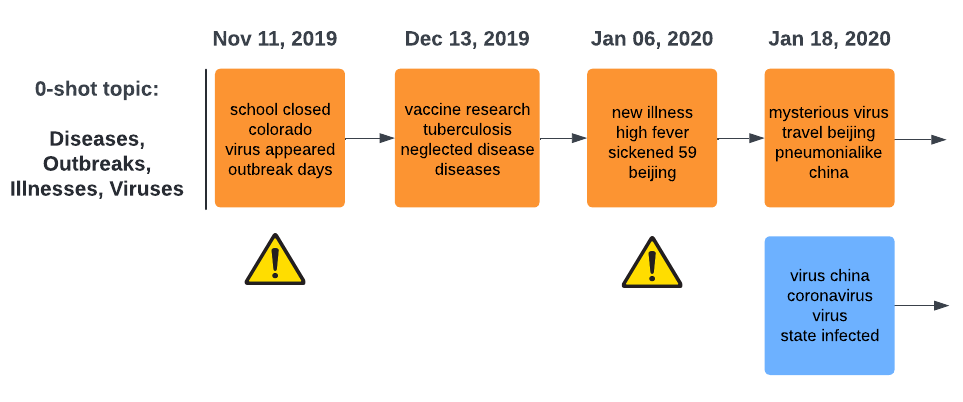}
    \caption{Comparison of COVID-19 Signal Detection with and without zero-shot Topic Modeling}
    \label{fig:zero-shot_comparison}
\end{figure}

Figure \ref{fig:zero-shot_comparison} illustrates the impact of incorporating zero-shot topic modeling in the BERTrend algorithm. In this approach, an expert defines a general topic of interest, and each document from a slice is compared against this topic using embedding similarity. Documents that surpass a certain similarity threshold are captured, allowing for targeted weak signal detection.
This method enables experts to focus on specific topics of interest while offering higher precision and sensitivity in weak signal detection. By performing document-level comparisons using embeddings, the zero-shot approach minimizes the risk of missing relevant documents during the topic modeling pipeline.

In the provided example, we chose the generic zero-shot topic \texttt{"Diseases, Outbreaks, Illnesses, Viruses,"} to detect the COVID-19 signal, simulating a scenario where an expert has a general idea of what to monitor but lacks precise knowledge of an impending outbreak. Remarkably, the zero-shot method identified the earliest article in the dataset mentioning the coronavirus pandemic on January 6th, 2020, referring to it as a "pneumonia-like mysterious virus" alongside "coronavirus". This detection occurred 12 days before the automatic BERTrend usage without zero-shot.
Furthermore, the zero-shot approach captured potential weak signals even earlier, such as a November 2019 article reporting school closures in Colorado due to a virus outbreak. While these signals may or may not be directly related to the pandemic, they demonstrate the method's ability to identify potentially relevant events.
The consistency of the signal's growth is also notable. 
The automatically detected signal (blue) by BERTrend starts to decrease and becomes less stable around March 2020, not due to a loss in popularity, but because other signals discussing slightly different aspects of the pandemic begin to emerge. 



\section{Interpretation of trends with LLMs}
Topic modeling methods often output topics as sets of keywords, which can be difficult to interpret and may not fully capture the semantic meaning of the topic \citep{rijcken2023towards, rudiger2022topic}.

\begin{figure*}[t]
    \centering
    \includegraphics[width=1\linewidth]{Figures/LLM_Interpretation.pdf}
    \caption{Enhancing Signal Interpretation and Analysis using LLMs}
    \label{fig:llm_interpretation}
\end{figure*}


















 {\color{Black}LLMs can be leveraged to enhance the interpretation of signals detected by BERTrend and of their evolution over time. Although this field of topic analysis through LLMs is new, it is quite promising \cite{kirilenko-2024}.
 
 In this work, we go several steps further by using LLMs not only for having human-readable descriptions of topics, but also useful insights about their evolution between two timestamps, such a summary of the key developments of the event signal since previous timestamp, as well as novelty about the signal w.r.t. previous time period. In addition, we use the LLM to obtain an in-depth analysis of the signal, including: (1) impact, i.e. potential effects of this signal on various sectors, industries, and societal aspects, with both short-term and long-term implications; (2) evolution scenarios - both optimistic and pessimistic scenarios; (3) potential interactions /conflicts  with other current trends; (4) drivers and inhibitors (factors/barriers related to the development of the signal. The associated prompt templates are provided in section \ref{sec:prompts}.
 }

In the example of Figure \ref{fig:llm_interpretation}, we use the {\color{Black}GPT-4o} model \footnote{\url{https://platform.openai.com/docs/models/gpt-4o}} with a temperature of 0.1 to generate insightful summaries and highlight new information at each timestamp for a weak signal related to the new Bluetongue viral disease (Catarrhal fever) affecting ruminants that appeared in France in July 2024. This example was selected for its recency to ensure it lies beyond the LLM's training data, minimizing the risk of analysis bias from the model's pre-existing knowledge..

{\color{Black}By emphasizing new information at each timestamp through a multi-faceted description, the LLM helps to pinpoint key developments and changes within the topic. It provides a comprehensive summary of the signal's evolution, which can then be reintroduced to the LLM for further analysis, assessing its potential impact and possible outcomes.}





\section{Conclusion}


In this paper, we introduced BERTrend, a novel framework for detecting and monitoring weak signals in large, evolving text corpora. BERTrend models the trends of topics over time and classifies them as weak signals, strong signals, or noise based on their popularity metric.
The classification is performed using empirically chosen thresholds based on the distribution of topic popularities over a sliding window.
The other contributions of this work include: (1) an extensive evaluation on two real-world datasets 
that demonstrate the effectiveness of our approach; (2) proposals to leverage LLMs to enhance the interpretation of topic evolution. 


{\color{Black}We are currently exploring LLM-generated evolving knowledge graphs as a structured method for interpreting signals. These graphs monitor topic evolution by tracking the appearance and disappearance of entities and relationships. Future work will involve exploring new datasets, integrating live data, and developing metrics to compare weak signal detection methods.}

{\color{black}
\section{Software availability}
In order to foster collaboration and advancement in weak signal detection, the code of BERTrend 
(and associated tools for visualization and LLM-based interpretation) has been open-sourced.
It is available at the following URL: 

\noindent\url{https://github.com/rte-france/BERTrend}.
}

\section{Limitations}
\label{sec:limitations}

%

\subsection{Hyperparameter Sensitivity}
BERTrend's performance is sensitive to various hyperparameters, including BERTopic parameters, merge threshold, granularity, and retrospective period. We chose BERTopic hyperparameters to produce the most fine-grained topics since larger topics will hinder the early detection process, and weak signals will get lost as the documents that should form them are assigned either to noise topics or other large, more generalized topics. To mitigate the variability of topic embeddings due to the small number of documents per topic, we selected a low merge threshold (0.6-0.7). Granularity depends on the amount of data available per time unit and the frequency of new documents. The retrospective period affects the influence of past signals on current thresholds; we found that a period of a week to a month doesn't change thresholds significantly, but bigger changes can affect classification results. Empirically fixed thresholds (10th percentile and median) balance precision and recall.


\subsection{Distinguishing Between Weak Signals and Noise}
There remains the challenge of distinguishing between what's considered a weak signal and what's considered noise. Relying on temporal popularity fluctuations alone isn't ideal, as both weak and noise signals behave very similarly. There's also the issue of characterizing what would be a "weak signal," since that changes from one person to another, one domain to another, etc. This is why we added the zero-shot detection to help an expert guide the detection process. We envision exploring the effect of using named entity recognition for better filtering in future work.


\subsection{Evaluation Challenges}
\label{sec:evaluation_challenges}
Evaluating the effectiveness of our weak signal detection method is challenging due to many factors:
\begin{itemize}[nolistsep, leftmargin=*]
    \item  the  subjective nature of what constitutes a weak signal, since it depends on the context, the domain, and the specific goals of the analysis, making it difficult to raise a consensus even among domain experts.
    \item the lack of ground truth data: unlike many other natural language processing tasks, there are no widely accepted benchmark datasets or ground truth annotations specifically designed for evaluating weak signal detection. This lack of standardized benchmarks hinders the ability to objectively compare different approaches and quantify their performance.
    \item dynamics over time: weak signals are often transient and can grow or dissipate over time. This dynamic nature complicates the evaluation process, as the ground truth itself may change, requiring continuous monitoring and updating of the evaluation data.
\end{itemize}

{\color{Black}
To the best of our knowledge, there are currently no established metrics for comparing weak signal detection performance 
within large volumes of data.
Traditional metrics used in evaluating topic models, such as topic coherence 
topic diversity, and perplexity, are not suitable for assessing weak signal detection. These metrics measure the quality and interpretability of topics over time, but they cannot determine whether a detected signal is truly a weak signal of emerging importance.
Given this context, comparing BERTrend with dynamic topic models or other embedding techniques (as described in \citet{balepur2023-dynamite}, \citet{churchill2022}, \citet{rudolph2018}, \citet{yao2018}, \citet{meng2020}, or \citet{xu2023}) using these metrics would not provide meaningful insights into the nature of the weak signals detected. These methods and their evaluation metrics are designed for different objectives, primarily assessing topic quality and evolution over extended periods of time.

Comparing BERTrend with existing keyword-based approaches (e.g., \citet{park2017future}; \citet{donnelly2019application}; \citet{griol2020detecting}) is not feasible due to fundamental differences in methodology and output:
(1) These methods primarily use Degree of Visibility and Degree of Diffusion metrics on keyword emergence maps and keyword issue maps. Their output is a set of words indicating the presence of a weak signal, whereas BERTrend produces topic sequences over time.
(2) BERTrend's dynamic, embedding-based approach captures contextual nuances that keyword-based methods often miss. As noted by \citet{rousseau2021}, "the use of a single keyword may lead to a loss of objectivity" and "the lack of relations and context over the keywords limit the information."

To address the evaluation challenge, our future work will center on a large-scale user study involving domain experts. These experts will review BERTrend's outputs at specific time instants, identifying potential weak signals in their fields. 
}




\bibliography{custom}















\appendix



\section{Software}
\label{sec:appendix_software}

\subsection{Some screenshots}
\begin{figure*}[h]
    \centering
    \includegraphics[width=1\linewidth]{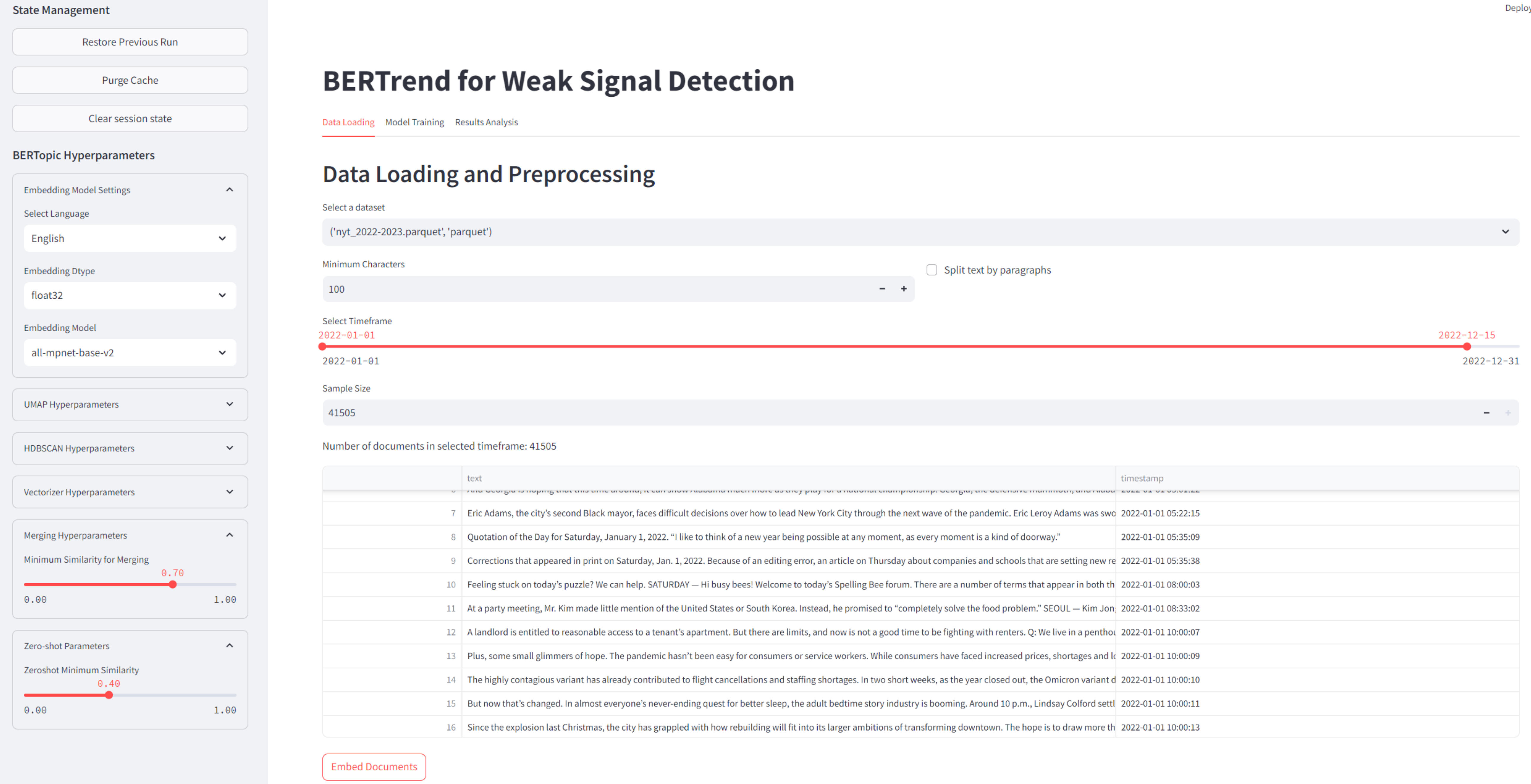}
    \captionsetup{justification=centering}
    \caption{The BERTrend main interface allows users to configure various hyperparameters, including those for BERTopic components and merging thresholds. Users can load and filter data, split text into paragraphs, select specific timeframes, and randomly sample the data. The interface also facilitates the embedding of documents for further analysis.}
    \label{fig:data-loader}
\end{figure*}

\begin{figure*}[h]
    \centering
    \includegraphics[width=1\linewidth]{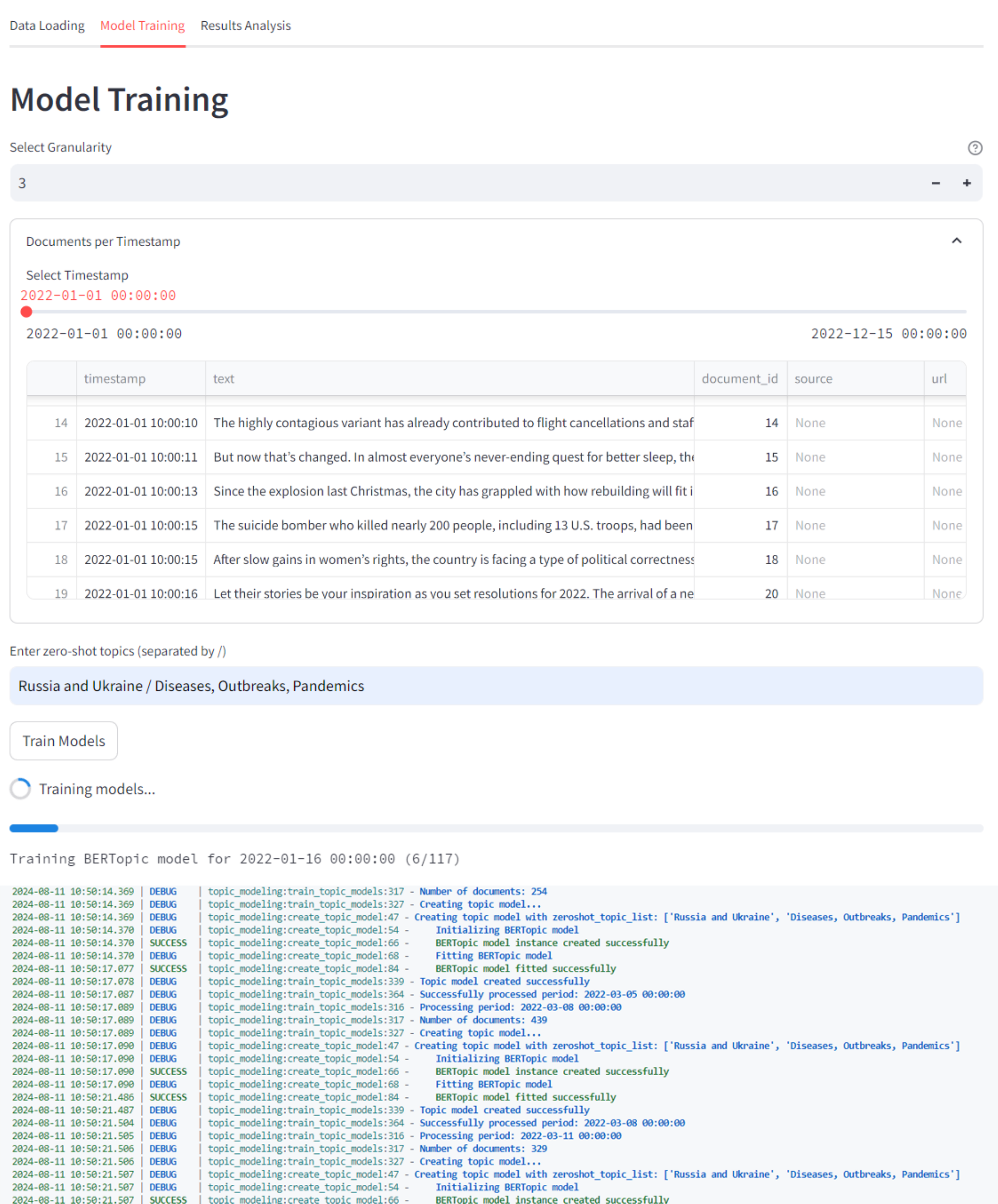}
    \captionsetup{justification=centering}
    \caption{The model training interface enables the creation and merging of multiple BERTopic models based on the selected granularity and merging thresholds. Users can also define zero-shot topics for detection at each timestamp, providing a flexible approach to model training.}
    \label{fig:model-training}
\end{figure*}

\begin{figure*}[h]
    \centering
    \includegraphics[width=1\linewidth]{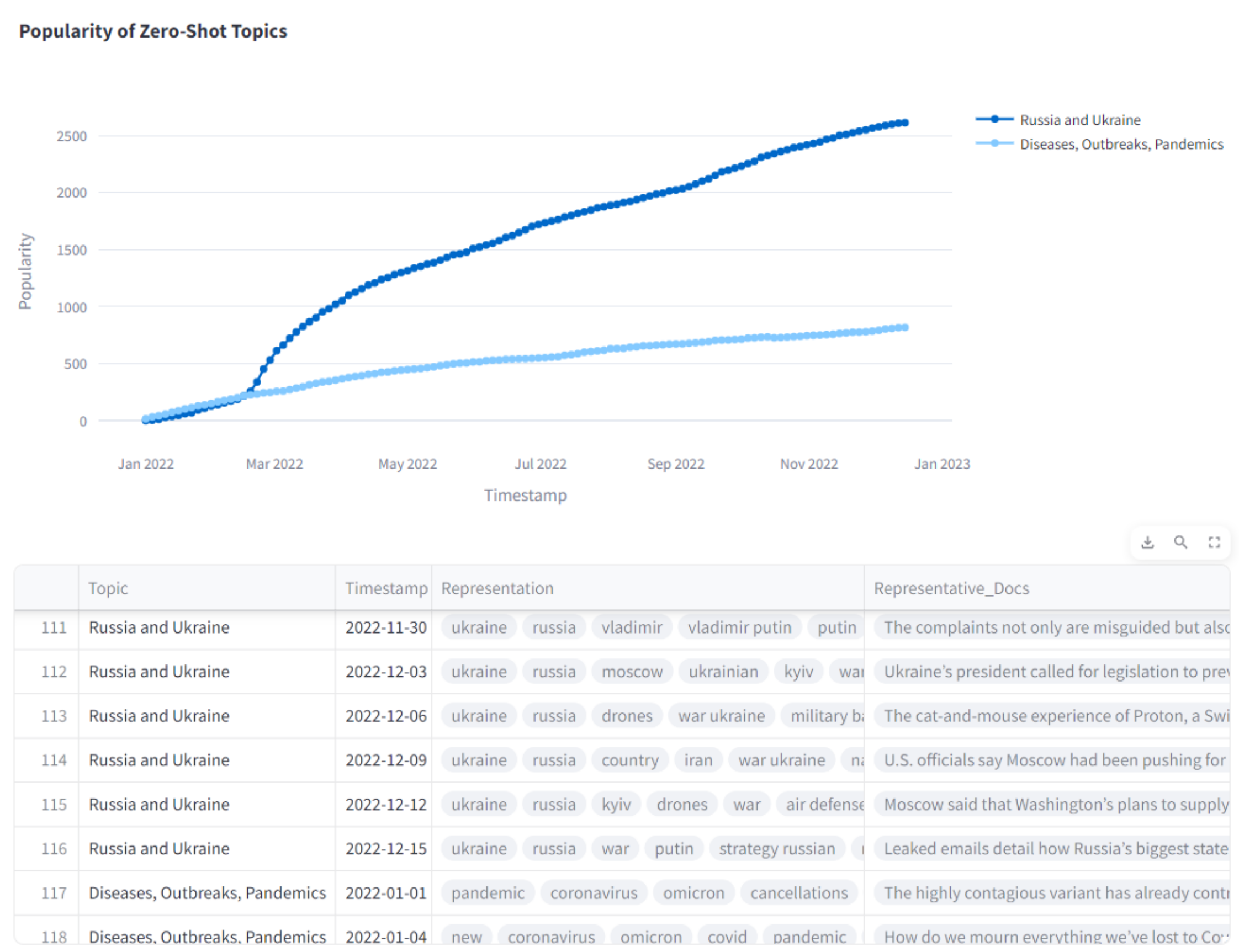}
    \captionsetup{justification=centering}
    \caption{The results page showcases zero-shot topics, allowing experts to visually inspect them with ease. A searchable dataframe accompanies the visualization, enabling users to explore documents related to defined zero-shot topics across various timestamps.}
    \label{fig:zeroshot-topics}
\end{figure*}

\begin{figure*}[h]
    \centering
    \includegraphics[width=1\linewidth]{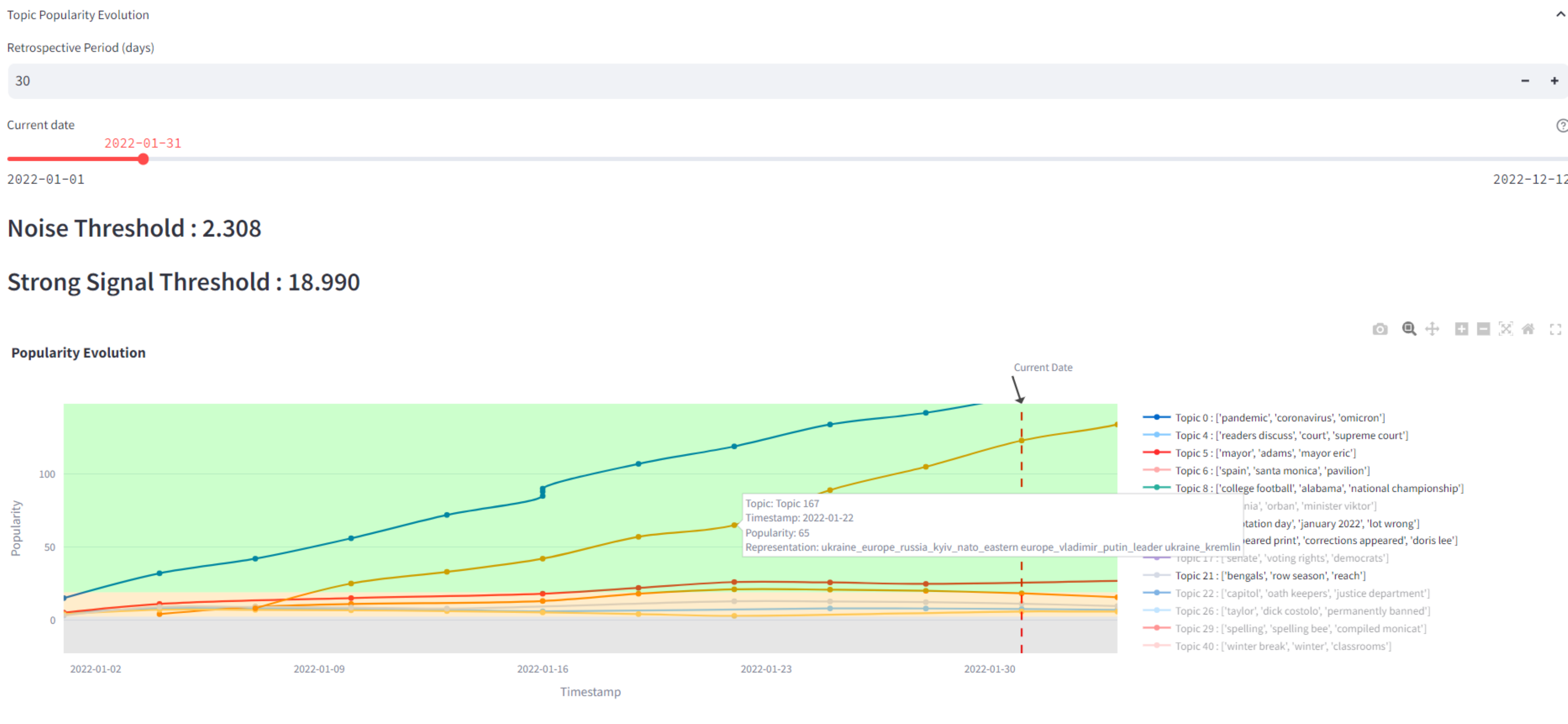}
    \captionsetup{justification=centering}
    \caption{The core functionality of BERTrend: users can define a retrospective period and select specific dates to investigate historical data, determining what was classified as noise, weak signals, or strong signals during that timeframe.}
    \label{fig:popularity-evolution}
\end{figure*}

\begin{figure*}[h]
    \centering
    \includegraphics[width=1\linewidth]{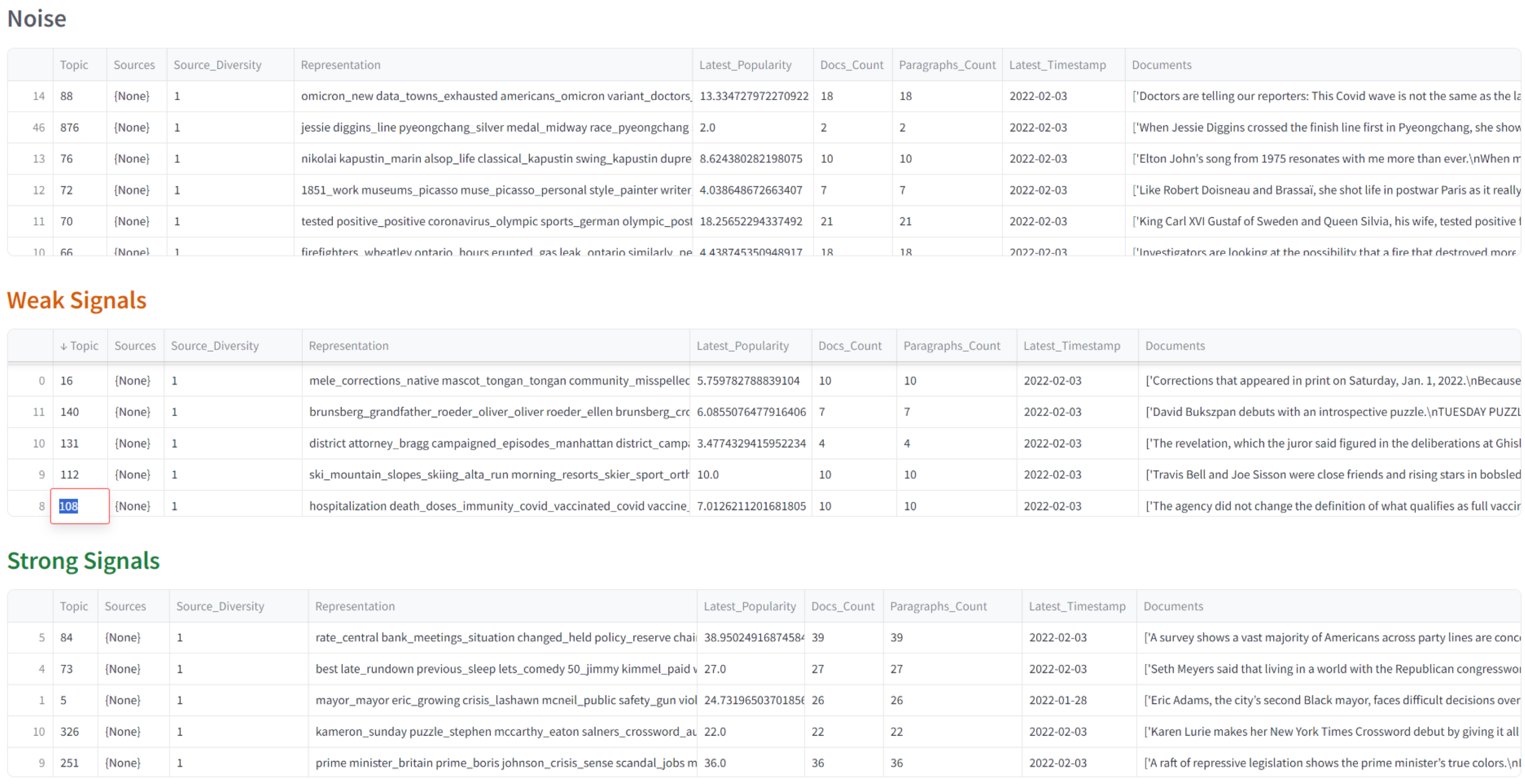}
    \captionsetup{justification=centering}
    \caption{For each selected date, corresponding dataframes classify topics based on their popularity, categorizing them as noise, weak signals, or strong signals. Users can easily retrieve and further analyze a topic by its identifier, as demonstrated with topic number 108.}
    \label{fig:signal-classification}
\end{figure*}

\begin{figure*}[h]
    \centering
    \includegraphics[width=1\linewidth]{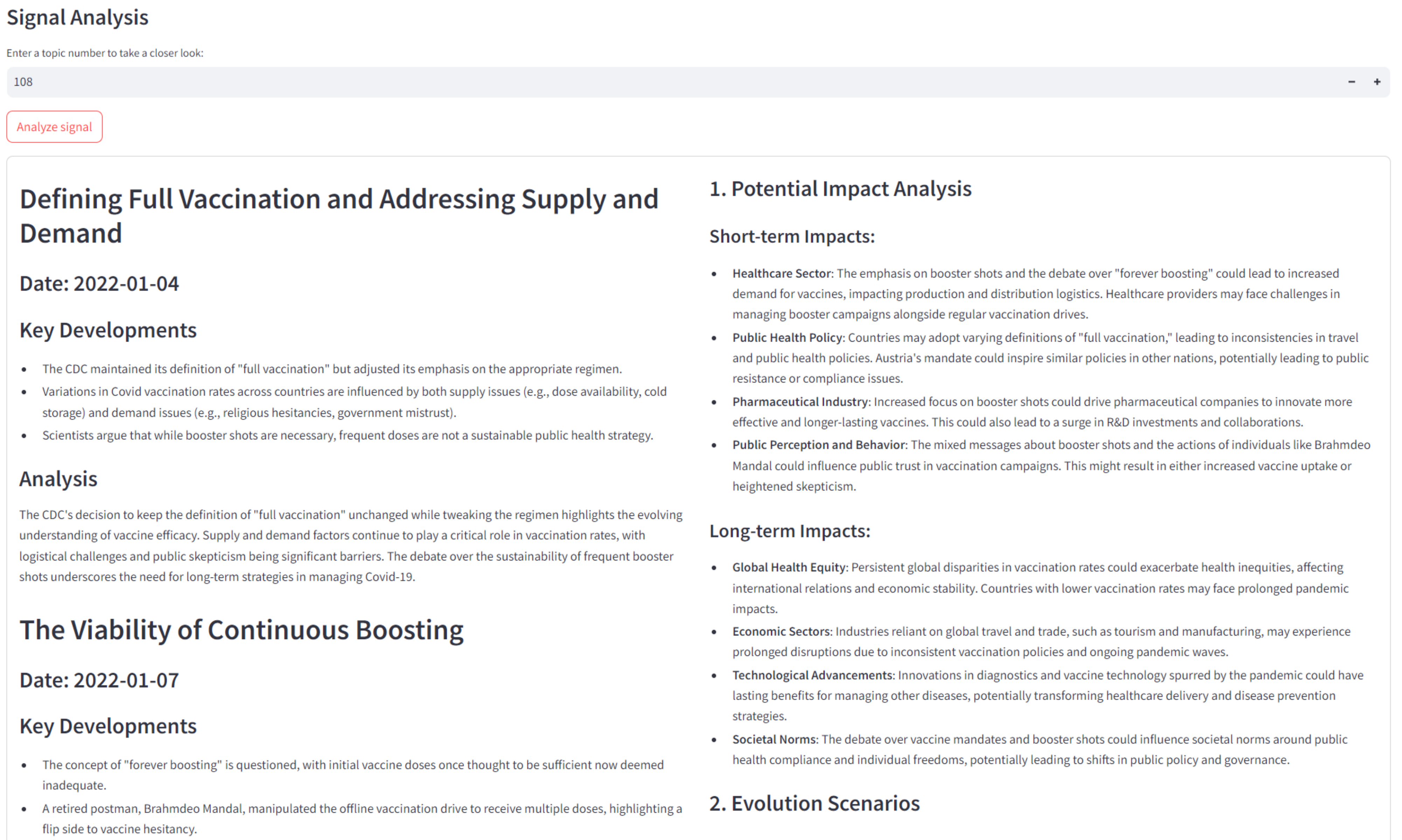}
    \captionsetup{justification=centering}
    \caption{Upon selecting a topic identifier, an LLM generates a comprehensive analysis of the topic's evolution and its various aspects, presented in a detailed report for further examination.}
    \label{fig:signal-analysis}
\end{figure*}

\begin{figure*}[h]
    \centering
    \includegraphics[width=1\linewidth]{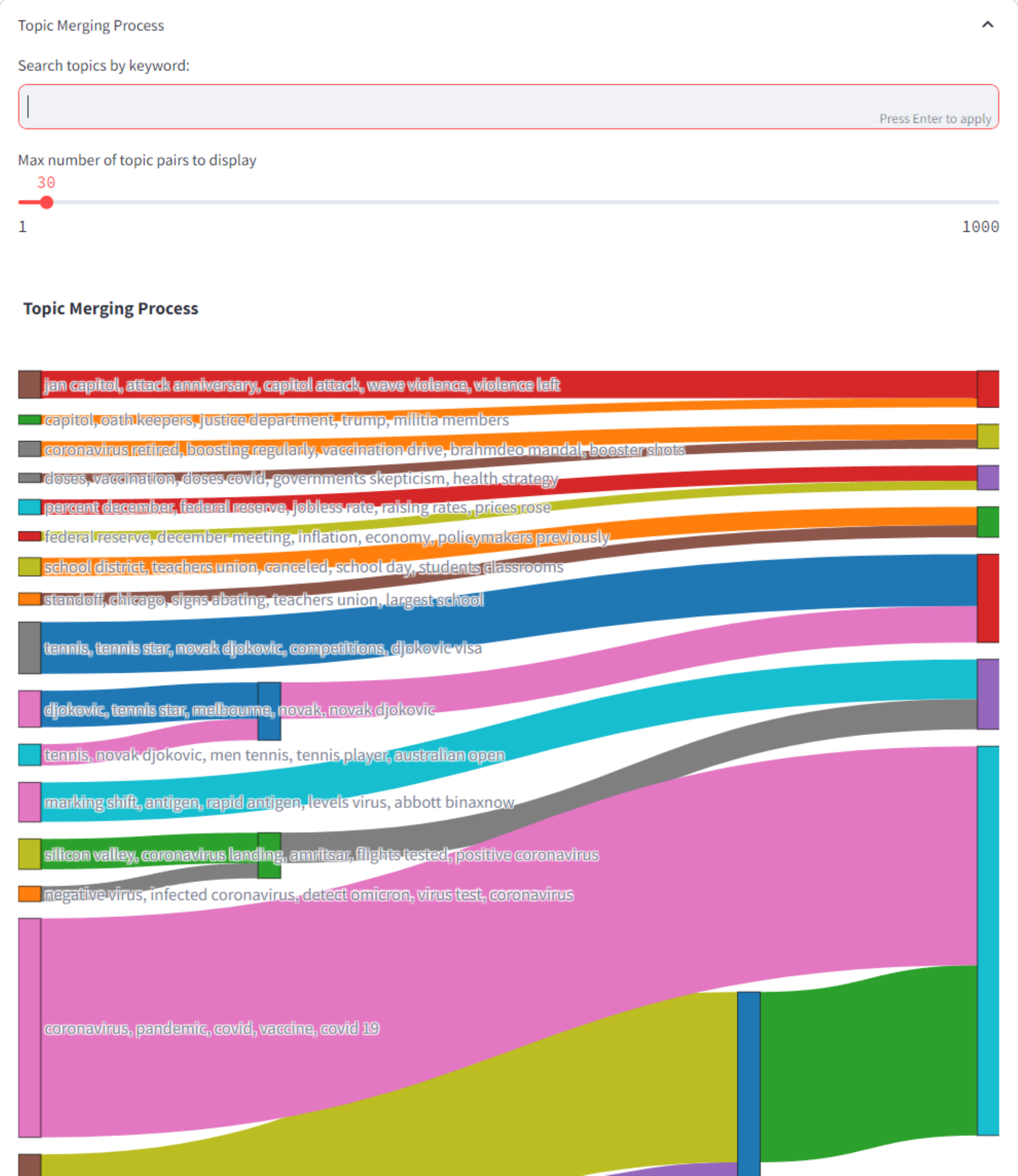}
    \captionsetup{justification=centering}
    \caption{The topic merging process is visualized using a Sankey Diagram, providing a clear and intuitive representation of how topics were combined over time.}
    \label{fig:sankey-diagram}
\end{figure*}

{\color{Black}
We present in this section some screenshots (Figures \ref{fig:data-loader}–\ref{fig:sankey-diagram}) of our prototype which utilizes BERTrend to explore trends and categorize them into different types of signals, as well as using a LLM to interpret and analyze certain signals. The UI is built using Streamlit\footnote{\url{https://streamlit.io/}}, and all the visualizations are done using the Plotly library\footnote{\url{https://plotly.com/}}.

}

{\color{Black}
\subsection{Prompt examples for topic evolution analysis}
\label{sec:prompts}
This section gives some examples of the prompts we are using with a LLM (GPT-4o) to obtain detailed insights of topic evolution between two timestamps.
}

{\color{Black}
\subsubsection{Prompt for evolving topic summary at a given timestamp}
{\small
\begin{verbatim}  
As an expert analyst specializing in trend analysis
and strategic foresight, your task is to provide a 
comprehensive evolution summary of Topic 
{topic_number}. Use only the information provided 
below:

{content_summary}

Structure your analysis as follows:

For the first timestamp:

## [Concise yet impactful title capturing the 
essence of the topic at this point]
### Date: [Relevant date or time frame]
### Key Developments
- [Bullet point summarizing a major development 
or trend]
- [Additional bullet points as needed]

### Analysis
[2-3 sentences providing deeper insights into the 
developments, their potential implications, and 
their significance in the broader context of the 
topic's evolution]

For all subsequent timestamps:

## [Concise yet impactful title capturing the 
essence of the topic at this point]
### Date: [Relevant date or time frame]
### Key Developments
- [Bullet point summarizing a major development
or trend]
- [Additional bullet points as needed]

### Analysis
[2-3 sentences providing deeper insights into the
developments, their potential implications, and 
their significance in the broader context of the 
topic's evolution]

### What's New
[1-2 sentences highlighting how this period differs
from the previous one, focusing on new elements or 
significant changes]

Provide your analysis using only this format, based 
solely on the information given. Do not include any 
additional summary or overview sections beyond what
is specified in this structure.

\end{verbatim}
}
}

\subsubsection{Prompt for signal analysis}
{\color{Black}
{\small
\begin{verbatim}  
As an elite strategic foresight analyst with 
extensive expertise across multiple domains and 
industries, your task is to conduct a comprehensive 
evaluation of a potential signal derived from the 
following topic summary:

{summary_from_first_prompt}

Leverage your knowledge and analytical skills to 
provide an in-depth analysis of this signal's 
potential impact and evolution:

1. Potential Impact Analysis:
   - Examine the potential effects of this signal 
   on various sectors, industries, and societal 
   aspects.
   - Consider both short-term and long-term 
   implications.
   - Analyze possible ripple effects and 
   second-order consequences.

2. Evolution Scenarios:
   - Describe potential ways this signal could 
   develop or manifest in the future.
   - Consider various factors that could influence
   its trajectory.
   - Explore both optimistic and pessimistic 
   scenarios.

3. Interconnections and Synergies:
   - Identify how this signal might interact with 
   other current trends or emerging phenomena.
   - Discuss potential synergies or conflicts with 
   existing systems or paradigms.

4. Drivers and Inhibitors:
   - Analyze factors that could accelerate or amplify
   this signal.
   - Examine potential barriers or resistances that 
   might hinder its development.

Your analysis should be thorough and nuanced, going 
beyond surface-level observations. Draw upon your 
expertise to provide insights that capture the 
complexity and potential significance of this signal. 
Don't hesitate to make well-reasoned predictions 
about its potential trajectory and impact.

Focus on providing a clear, insightful, and 
actionable analysis that can inform strategic 
decision-making and future planning.
\end{verbatim}
}
}

\end{document}